\documentclass[letterpaper, 10 pt, conference]{ieeeconf}
\IEEEoverridecommandlockouts

\usepackage[colorlinks]{hyperref}
\hypersetup{
    colorlinks,
    linkcolor=black,
    citecolor=black,
    filecolor=magenta,
    urlcolor=cyan,
}

\usepackage{resizegather}

\usepackage{cite}
\usepackage{amsmath,amssymb,amsfonts,mathrsfs}
\usepackage{mdframed}
\usepackage{graphicx}
\usepackage{textcomp}
\usepackage{xcolor}
\usepackage{tcolorbox}
\usepackage{textcomp}
\usepackage{tablefootnote}
\usepackage{threeparttable}
\usepackage{bbm}
\usepackage{dsfont}
\usepackage{tikz}
\usepackage{graphicx}
\usepackage{tkz-euclide}
\usepackage{algorithm}
\usepackage{algpseudocode}
\usepackage{enumerate}
\usepackage{mathtools}
\usepackage{balance}
\usepackage{flushend}

\newcommand{\vertiii}[1]{{\left\vert\kern-0.25ex\left\vert\kern-0.25ex\left\vert #1 
    \right\vert\kern-0.25ex\right\vert\kern-0.25ex\right\vert}}

\newtheorem{theorem}{\textbf{Theorem}}

\newtheorem{problem}{\textbf{Problem}}

\newtheorem{remark}{Remark}

\newtheorem{definition}{\textbf{Definition}}

\makeatletter
\renewcommand{\fps@figure}{htp}
\renewcommand{\fps@table}{htp}
\makeatother

\def\BibTeX{{\rm B\kern-.05em{\sc i\kern-.025em b}\kern-.08em
    T\kern-.1667em\lower.7ex\hbox{E}\kern-.125emX}}
    
\title{Neural Control Barrier Functions from Physics Informed Neural Networks} 

\author{Shreenabh Agrawal\(^{1, 2}\), Manan Tayal\(^{1}\), Aditya Singh\(^{1}\), and Shishir Kolathaya\(^{1, 3}\)
\thanks{This work was supported by the ARTPARK}
\thanks{\(^{1}\)Center for Cyber-Physical Systems, Indian Institute of Science (IISc), Bengaluru.}
\thanks{\(^{2}\)Department of Physics, IISc, Bengaluru.}
\thanks{\(^{3}\)Department of Computer Science and Automation, IISc, Bengaluru.}
\thanks{\tt \{shreenabhm, manantayal, adityasingh, shishirk\}
@iisc.ac.in}
}

\begin{document}

\maketitle

\begin{abstract}
    As autonomous systems become increasingly prevalent in daily life, ensuring their safety is paramount. Control Barrier Functions (CBFs) have emerged as an effective tool for guaranteeing safety; however, manually designing them for specific applications remains a significant challenge. With the advent of deep learning techniques, recent research has explored synthesizing CBFs using neural networks—commonly referred to as neural CBFs. This paper introduces a novel class of neural CBFs that leverages a physics-inspired neural network framework by incorporating Zubov's Partial Differential Equation (PDE) within the context of safety. This approach provides a scalable methodology for synthesizing neural CBFs applicable to high-dimensional systems. Furthermore, by utilizing reciprocal CBFs instead of zeroing CBFs, the proposed framework allows for the specification of flexible, user-defined safe regions. To validate the effectiveness of the approach, we present case studies on three different systems: an inverted pendulum, autonomous ground navigation, and aerial navigation in obstacle-laden environments.
\end{abstract}

\section{Introduction}
\label{section: Introduction}
Ensuring safety is a primary challenge in control systems, especially in autonomous applications where real-time guarantees are essential. Any safety violation can have severe consequences; therefore, designing control strategies that both safeguard operations and maintain efficiency is imperative. 

To address this challenge, several paradigms have been developed. Constrained Reinforcement Learning (CRL) methods \cite{altman2021constrained,achiam2017constrained} embed safety constraints within a reinforcement learning framework to learn safe policies in a data-driven manner. However, due to their reliance on exploration, these methods may not consistently enforce strict safety guarantees. Alternatively, Hamilton-Jacobi (HJ) reachability analysis \cite{8263977,tayal2025physics,singh2024exactimpositionsafetyboundary} provides rigorous worst-case assurances by computing backward-reachable sets that delineate a safe operating domain. Despite its theoretical robustness, the substantial computational burden of HJ reachability makes it less practical for real-time, high-dimensional systems.

A particularly promising approach involves the use of Control Barrier Functions (CBFs) \cite{ames2014control}, which offer an effective means of synthesizing safe control laws for control affine systems \cite{Ames_2017}. By formulating the control design as Quadratic Programs (QPs), which can be solved at high frequencies using modern optimization solvers, CBFs have been successfully applied to various safety-critical tasks such as adaptive cruise control \cite{ames2014control}, aerial maneuvers \cite{7525253,tayal2024control}, and legged locomotion \cite{ames2019control,C3BF-Legged}. Nonetheless, the safety and performance of these applications are highly dependent on the chosen CBF, and traditional synthesis methods—such as sum-of-squares techniques \cite{1470374,TOPCU20082669}—often struggle with high-dimensional state spaces.

In recent years, the universal approximation capabilities of neural networks have spurred interest in neural network-based barrier function (NCBF) synthesis. A diverse range of approaches has emerged, including methods that leverage expert demonstrations \cite{9303785}, SMT-based techniques \cite{zhao2020synthesizing,abate2021fossil}, mixed-integer programming \cite{zhao2022verifying}, and nonlinear programming \cite{NEURIPS2023_120ed726}. Recent work has also introduced specialized loss functions for training NCBFs \cite{dawson2023safe,liu2023safe,tayal2025cp,9993334}. However, while methods that produce formally verified CBFs \cite{tayal2024learning, tayal2024semi} offer strong guarantees, they often lack scalability when applied to high-dimensional systems. Moreover, none of the previously discussed approaches provide the flexibility to select a user-defined sub-level set, which is essential for tailoring the safe region to particular requirements.

Building on recent advances in Neural Control Lyapunov Function (NCLF) synthesis using Physics Informed Neural Networks (PINNs) with Zubov's method \cite{clf-zubov,liu2023learningverifyingmaximalneural,liu2024formallyverifiedphysicsinformedneural,liu2025physicsinformedneuralnetworklyapunov}, this paper systematically investigates a novel framework for learning Neural Control Barrier Functions (NCBFs) through a Zubov's PDE based construction. Incorporating Zubov's PDE during training leads to the development of maximal neural Lyapunov functions with verifiable domains of attraction that closely approximate the analytical solutions \cite{liu2023learningverifyingmaximalneural,liu2024formallyverifiedphysicsinformedneural}. 

Prior studies in stability analysis have demonstrated that Zubov’s theorem provides a useful characterization of domains of attraction, by transforming the Lyapunov function learning problem into a bounded PDE problem. By constructing a transformed Lyapunov function that is guaranteed to be bounded, this method overcomes the numerical issues encountered in the synthesis, improves training stability, and allows users to adjust sub-level sets according to their requirements. Recognizing that similar challenges exist in the synthesis of Reciprocal Control Barrier Functions (RCBFs), our work introduces a PINN-based methodology to learn them, thereby integrating the advantages of Zubov's approach into the training process. 

The primary contributions of this paper are as follows:  

\begin{itemize}
    \item We reformulate Zubov’s PDE in the context of safety to synthesize neural CBFs using PINNs and design a safe controller through its QP formulation.  
    \item Our framework allows for flexible level set selection, enabling users to tailor the safe region to specific requirements.  
    \item We achieve efficient and reliable training, demonstrating rapid convergence and high sample efficiency with only 10,000 random samples and 2,000 training epochs across diverse environments.  
    \item To validate our approach, we conduct experiments on an inverted pendulum, autonomous ground navigation, and aerial navigation in obstacle-rich environments, showcasing scalability to high-dimensional systems.  
\end{itemize}

\section{Preliminaries}
\label{section: Background}
In this section, we formally introduce Control Barrier Functions (CBFs) and their importance for real-time safety-critical control. We consider a continuous-time deterministic control system with state \(x(t)\in D\subseteq\mathbb{R}^n\) and control input \(u(t)\in\mathbb{U}\subseteq\mathbb{R}^m\) for \(t\geq 0\). The state dynamics are described by the ordinary differential equation
\begin{equation}
\label{eq:control_system_deterministic}
\dot{x}(t) = f(x(t)) + g(x(t))u(t),
\end{equation}
where \(f:\mathbb{R}^{n}\rightarrow\mathbb{R}^{n}\) and \(g:\mathbb{R}^{n}\rightarrow\mathbb{R}^{n\times m}\) are continuously differentiable functions, and \(u(t)\in\mathbb{R}^{m}\) is the control input. Moreover, given a Lipschitz continuous control law \(u=k(x)\), the closed-loop system is expressed as
\begin{equation}
    \label{eq: Closed loop control}
    \dot{x}(t)=f(x(t))+g(x(t))k(x(t)),
\end{equation}

For any initial condition \(x(0)=x_{0}\), we denote the unique solution by \(\phi(t,x_{0})\) for \(t\) in the maximal interval of existence \(J\). 

\begin{definition}[Safe Set]
A set \(\operatorname{Int}(\mathcal{C})\subseteq \mathbb{R}^{n}\) is said to be a \emph{safe set} for the closed-loop system under the control law \(u=k(x)\) if for every initial state \(x(0)\in\operatorname{Int}(\mathcal{C})\), the unique solution \(\phi(t,x(0))\) satisfies
\begin{equation}
    \label{eq:safety_condition_invariance}
    \phi(t,x(0))\in\operatorname{Int}(\mathcal{C})\quad \text{for all } t\geq 0.
\end{equation}
\end{definition}

This property ensures that once the state is within \(\mathcal{C}\), it remains safe for all future times. The distance of a point \(x\in\mathbb{R}^{n}\) to a set \(\mathcal{C}\subseteq\mathbb{R}^{n}\) is defined as
\(
\|x\|_{\mathcal{C}}:=\inf_{y\in\mathcal{C}}\|x-y\|.
\)

\subsection{Reciprocal Control Barrier Functions (RCBFs)}
\label{subsec:rbf}

In line with the definition of safe set above, we now represent \(\operatorname{Int}(\mathcal{C})\) equivalently as the super-level set of a continuously differentiable function \(h: D\to\mathbb{R}\). That is, we define
\begin{align}
\label{eq:setc1}
    \mathcal{C} &= \{ x \in D \subset \mathbb{R}^n : h(x) \geq 0\}, \\
\label{eq:setc2}
    \partial\mathcal{C} &= \{ x \in D \subset \mathbb{R}^n : h(x) = 0\}, \\
\label{eq:setc3}
    \operatorname{Int}(\mathcal{C}) &= \{ x \in D \subset \mathbb{R}^n : h(x) > 0\}.
\end{align}
It is assumed that \(\operatorname{Int}(\mathcal{C})\neq \emptyset\) and that \(\mathcal{C}\) has no isolated points, i.e., \(\overline{\operatorname{Int}(\mathcal{C})} = \mathcal{C}\). We then define the reciprocal barrier function as
\begin{equation}
    \label{eq: RBF definition}
    B(x)=\frac{1}{h(x)}.
\end{equation}

We can verify that the controller \(k(x)\) ensures safety by using the notion of Reciprocal Control Barrier Functions (RCBFs), which are defined next.

\begin{definition}[Reciprocal Control Barrier Function]
A continuously differentiable function \(B: \operatorname{Int}(\mathcal{C}) \rightarrow \mathbb{R}\) is called a \emph{reciprocal control barrier function (RCBF)} if it is defined by \eqref{eq: RBF definition} and satisfies:
\begin{enumerate}
    \item \textbf{Barrier Property:} \(B(x)\) is finite for all \(x\in \operatorname{Int}(\mathcal{C})\). Moreover, since \(h(x)\rightarrow 0\) as \(x\rightarrow \partial\mathcal{C}\), we have
    \begin{equation}
        \label{eq: RBF blowup}
            B(x)=\frac{1}{h(x)}\rightarrow \infty \quad \text{as } x\rightarrow \partial\mathcal{C},
    \end{equation}
    ensuring that unsafe conditions (i.e., approaching the boundary of \(\mathcal{C}\)) are heavily penalized.
    \item \textbf{Descent Along Trajectories:} The time derivative along trajectories,
    \begin{equation}
        \label{eq: RBF time derivative}
            \dot{B}(x):=\lim_{t\to 0^+}\frac{B(\phi(t,x))-B(x)}{t},
    \end{equation}
    is well-defined and satisfies
    \begin{equation}
        \label{eq: RBF derivative}
            \dot{B}(x)=-\Phi(x),
    \end{equation}
    where \(\Phi: D\to \mathbb{R}\) is continuous and positive.
\end{enumerate}
\end{definition}

\begin{definition}[Safety Conditions for RCBF]
\label{def: safety_conditions}
Let \(B\) be an RCBF as in the previous definition. Then, \(B\) is said to satisfy the \emph{safety conditions} if there exist class \(\mathcal{K}\) functions \(K_1\), \(K_2\), and \(K_3\) such that for all \(x \in \operatorname{Int}(\mathcal{C})\):
\begin{enumerate}
    \item[1)] \textbf{Boundedness Condition:} {This condition asserts that the barrier function \(B(x)\) is bounded above and below by class \(\mathcal{K}\) functions of \(h(x)\) as follows:}
    \begin{equation}
        \label{eq: RCBF bounded}
        \frac{1}{K_1(h(x))} \leq B(x) \leq \frac{1}{K_2(h(x))}.
    \end{equation}
    \item[2)] \textbf{Safety Condition:} {Enforcing safety requires that the following inequality is satisfied:}
    \begin{equation}
        \label{eq:rbf_condition}
        \inf_{u\in\mathbb{U}} \Bigl[ L_fB(x) + L_gB(x) u \Bigr] \le K_3\Bigl(\frac{1}{B(x)}\Bigr),
    \end{equation}
    where the Lie derivatives are given by
    \begin{equation}
        \label{eq: lie}
        L_fB(x)=\nabla B(x)^\top f(x), \quad L_gB(x)=\nabla B(x)^\top g(x).
    \end{equation}
    \item[3)] \textbf{Control Set Definition:} {Accordingly, we define the set of admissible control inputs as (using \eqref{eq: RBF definition} to simplify)}
    \begin{equation}
        \label{eq: Krcbf}
        K_{\mathrm{rcbf}}(x)=\left\{u \in \mathbb{U} \,\middle|\, L_fB(x) + L_gB(x) u \le K_3(h(x))\right\}.
    \end{equation}
\end{enumerate}
\end{definition}

If a controller selects inputs \(u\) from \(K_{\mathrm{rcbf}}(x)\) for all \(x \in\operatorname{Int}(\mathcal{C})\), then the safe set is forward invariant \cite{Ames_2017, ames2019control}. The RCBF \(B\) is said to be locally Lipschitz continuous if \(K_3\) and \(\frac{\partial B}{\partial x}\) are both locally Lipschitz continuous.

\subsection{Controller Synthesis}
\label{subsec:safe_controller}

Having established the barrier condition in \eqref{eq:rbf_condition}, we now design a control policy that minimally modifies a nominal control input \(u_{ref}(x)\) to ensure safety. In particular, the nominal controller is adjusted through a Quadratic Program (QP) that enforces the barrier condition as a constraint. The QP formulation is given by
\begin{equation}
\label{eq:rbf_qp}
\begin{aligned}
u^{*}(x) =\; &\underset{u\in\mathbb{U}\subseteq\mathbb{R}^m}{\text{argmin}}\; \|u-u_{ref}(x)\|^2 \\
& \textrm{s.t. } \; L_fB(x)+L_gB(x)u - \kappa h(x)\leq 0,
\end{aligned}
\end{equation}
where \(\mathbb{U}\subseteq\mathbb{R}^m\) denotes the set of admissible control inputs and \(\kappa\) is a positive constant (by choosing $K_3(s) = \kappa s$).

Alternatively, when the safety condition (Eq. \ref{eq:rbf_condition}) is violated, a closed-form minimal-norm adjustment can be computed as
\begin{equation}
    \label{eq: delta u}
    \Delta u =
\begin{cases}
-\dfrac{L_fB(x)+L_gB(x)u_{ref}(x)-\kappa h(x)}{\|L_gB(x)\|^2+\varepsilon}\,L_gB(x), \\ \text{if } L_fB(x)+L_gB(x)u_{ref}(x) > \kappa h(x),\\[2mm]
0,   \text{otherwise},
\end{cases}
\end{equation}

where \(\varepsilon>0\) is a small regularization constant ensuring the expression is well-defined.

The final control law is then defined as
\begin{equation}
    \label{eq: final control}
    u(x) = u_{ref}(x) + \Delta u.
\end{equation}

This controller modifies the nominal input \(u_{ref}(x)\) only when necessary to ensure that the condition \eqref{eq:rbf_condition} holds for all \(x\) in \(\operatorname{Int}(\mathcal{C})\). Under standard assumptions, this guarantee implies that the safe set remains forward invariant under the deterministic dynamics.

\section{Problem Formulation}
\label{section: problem}
As described in Section~\ref{section: Introduction}, we now present an equivalent Zubov-type theorem for safe regions, drawing parallels from the work on Lyapunov functions and domains of attraction \cite{liu2025physicsinformedneuralnetworklyapunov,liu2023learningverifyingmaximalneural}.

\subsection{Zubov's Characterization for Safe Regions}

Define a function \(W_N: D \to \mathbb{R}\) that is related to \(B\) via a scalar transformation (made explicit later in \eqref{eq: W in terms of B}, \eqref{eq: W in terms of B explicit}). The following result holds.

\begin{theorem}[Zubov's Characterization for Safe Regions]
    {
    \textit{Let \(D\subseteq\mathbb{R}^n\) be an open set containing $\mathcal{C}$. Then, the forward invariant safe set is characterized by}
\begin{equation}
    \label{eq: Zubov safety condition}
    \operatorname{Int}(\mathcal{C}) = \{x \in D : W_N(x) < 1\},
\end{equation}
\textit{if and only if there exist continuous functions \(W_N: D\to\mathbb{R}\) and \(\Psi: D\to\mathbb{R}\) such that:}
\begin{enumerate}
    \item \textit{\(W_N(x)<1 \,\forall\, x \in \operatorname{Int}(\mathcal{C}), \,W_N(x)\rightarrow 1\) as \(x\rightarrow \partial \mathcal{C}\) and  \(W_N(x)=1 \,\forall\, x\in D\setminus\mathcal{C}\).
    \item \(\Psi\) is positive definite with respect to \(\mathcal{C}\).}
    \item \textit{For any sufficiently small \(c_3>0\), there exist constants \(c_1,c_2>0\) such that $\|x\|_{\mathcal{C}}\ge c_3 \Longrightarrow W_N(x)>c_1$ and  $\Psi(x)>c_2$.}

    \item \textit{\(W_N\) and \(\Psi\) satisfy the following PDE:}
    \begin{equation}
        \label{eq: Zubov PDE}
            \dot{W}_N(x) = -\Psi(x)\bigl(1-W_N(x)\bigr),
    \end{equation}

    \textit{where \(\dot{W}_N(x)\) denotes the derivative of \(W_N\) along trajectories of \eqref{eq:control_system_deterministic}}.
\end{enumerate}
    }
\end{theorem}

This theorem is closely related to the satisfaction of the conditions in Section~\ref{subsec:rbf}, and follows directly from Zubov’s theorem \cite{zubov1964methods} for domains of attraction (characterized by sub-level sets of Lyapunov functions). By a reciprocal relationship (see \cite{converse}), a Lyapunov function can be transformed into an RBF, thereby preserving the forward-invariant sub-level set structure, which defines the safe set in this case.

\subsection{Construction of \texorpdfstring{$W_N$}{WN}}

Let \(\beta:[0,\infty)\to\mathbb{R}\) satisfy:
\begin{equation}
    \label{eq: beta pde}
    \dot{\beta}(s)=(1-\beta(s))\,\psi(\beta(s)),\quad \beta(0)=0,
\end{equation}

where \(\psi:[0,\infty)\to\mathbb{R}\) is locally Lipschitz with \(\psi(s)>0\) for all \(s>0\). (Then \(\beta\) is continuously differentiable, strictly increasing, with \(\beta(s)\to 1\) as \(s\to\infty\).)

Assume that \(B:\mathbb{R}^n\to[0,\infty)\) is an RCBF candidate for \(\operatorname{Int}(\mathcal{C})\). Define the transformed function \(W_N:\mathbb{R}^n\to\mathbb{R}\) as:
\begin{equation}
    \label{eq: W in terms of B}
        W_N(x)=
    \begin{cases}
    \beta\bigl(B(x)\bigr), & \text{if } B(x)<\infty,\\[1mm]
    1, & \text{otherwise}.
    \end{cases}
\end{equation}

By construction, we have that \(W_N(x)<1\) for \(x\in \operatorname{Int}(\mathcal{C})\) and \(W_N(x)\to 1\) as \(B(x)\to\infty\) (i.e., as \(x\) approaches \(\partial\mathcal{C}\)).

To study the evolution of \(W_N\) along {trajectories} of \eqref{eq:control_system_deterministic}, we already know \eqref{eq: RBF derivative} from Section \ref{subsec:rbf}. Then, by the chain rule,
\begin{equation}
    \label{eq: W_N chain rule}
    \dot{W}_N(x)=\dot{\beta}\bigl(B(x)\bigr)\,\dot{B}(x)
=-\dot{\beta}\bigl(B(x)\bigr)\,\Phi(x).
\end{equation}

A particularly useful choice is to take \(\psi(s)=\alpha(1+s)\) for some constant \(\alpha>0\). In this case, \eqref{eq: beta pde} becomes
\begin{equation}
    \label{eq: beta pde solvable}
    \dot{\beta}(s)=\alpha\,(1-\beta(s))(1+\beta(s)),\quad \beta(0)=0.
\end{equation}

It can be verified that the unique solution is \(\beta(s)=\tanh(\alpha s)\). Thus, for \(B(x)<\infty\) we have
\begin{equation}
    \label{eq: W in terms of B explicit}
    W_N(x)=\tanh\bigl(\alpha\,B(x)\bigr).
\end{equation}

Differentiating along trajectories of \eqref{eq:control_system_deterministic} and using \eqref{eq: RBF derivative}, it follows that:
\begin{equation}
\label{eq: W simplification}
\begin{aligned}
\dot{W}_N(x) &= \alpha\Bigl[1-\tanh^2\bigl(\alpha\,B(x)\bigr)\Bigr]\dot{B}(x) \\
             &= \alpha\,(1-W_N(x))(1+W_N(x))\,\dot{B}(x) \\
             &= -\alpha\,(1-W_N(x))(1+W_N(x))\,\Phi(x).
\end{aligned}
\end{equation}

Thus, comparing with \eqref{eq: Zubov PDE}, we have:
\begin{equation}
    \label{eq: capital Psi}
    \Psi(x)=\alpha(1+W_N(x))\,\Phi(x).
\end{equation}

\medskip

\begin{remark}
    Two common choices for the transformation arise from different selections of \(\psi(s)\):
\begin{itemize}
    \item[(i)] If \(\psi(s)=\alpha\) (a constant), then \(\beta(s)=1-\exp(-\alpha s)\) is the solution of:
    \[
    \dot{\beta}(s)=\alpha(1-\beta(s)),\quad \beta(0)=0.
    \]
    \item[(ii)] For the numerical examples in this paper, we choose $\Phi(x) = ||x||^2_\mathcal{C}$ and \(\psi(s)=\alpha(1+s)\), then, as shown above, \(\beta(s)=\tanh(\alpha s)\).
\end{itemize}
In either case, the transformation yields a bounded function \(W_N\) that satisfies \(W_N(x)\to 1\) as \(x\) approaches the boundary of the safe set. In our analysis, we focus on the case (ii), which leads to the hyperbolic tangent transform.
\end{remark}

Having established the construction of \(W_N\), we now shift our focus to its approximation over the domain \(D\), thereby laying the groundwork for characterizing the safe region in practical settings. In particular, we pose the following problem:
 
\begin{problem}
 Given the control system in \eqref{eq:control_system_deterministic} and under the Zubov-type framework for safe regions (see \eqref{eq: Zubov safety condition} and \eqref{eq: Zubov PDE}), the goal is to devise an algorithm that learns the transformed reciprocal barrier function \(W_N(x)\) over \(D\) such that:
\begin{enumerate}
    \item \(W_N(x)\) is accurately approximated so that the safe set is characterized as in \eqref{eq: Zubov safety condition}.
    \item It can be used to construct a QP-based controller that ensures system safety.
\end{enumerate}
\end{problem}

The proposed approach benefits from the advantages of the Zubov-type construction over traditional RBFs, which typically diverge at the boundary. Because $W_N$ remains bounded and numerically more stable near the boundary ($\partial\mathcal{C}$), it is more convenient for practical learning methods.

\section{Methodology}
\subsection{Neural Network Training for the Reciprocal Barrier Function}
\label{subsec: NN training}

We propose the following neural network training scheme to solve the above-described problem statement. 

A physics-informed neural network (PINN) is employed to approximate the transformed reciprocal barrier function \(W_N(x;\theta)\) that satisfies the Zubov-type PDE given in \eqref{eq: Zubov PDE}. We can write \eqref{eq: Zubov PDE} as a general first-order PDE
\begin{equation}
    \label{eq: general_pde}
    F(x,W,\nabla W)=0, \quad x\in D,
\end{equation}
subject to the boundary condition
\begin{equation}
    \label{eq: bc}
    W_N(x;\theta)=\text{bc}(x),\quad x\in\partial D.
\end{equation}
We define \(\text{bc}(x)\) consistently with our construction as
\begin{equation}
    \label{eq: bc definition}
    \text{bc}(x)=
\begin{cases}
1, & \text{for } x\in\partial D\setminus \mathcal{C},\\[1mm]
\beta\bigl(B(x)\bigr), & \text{for } x\in\partial D\cap\mathcal{C},
\end{cases}
\end{equation}
with \(\beta\) defined as in \eqref{eq: beta pde} and \(B(x)\) being the reciprocal barrier function candidate.

To train \(W_N(x;\theta)\) as an approximation of the solution to \eqref{eq: general_pde}, we design a loss function comprising four components:
\begin{enumerate}
    \item \textbf{PDE Residual Loss:}  
    At a set of interior collocation points \(\{x_i\}_{i=1}^{N_c}\subset D\), we evaluate the residual using \eqref{eq: general_pde}:
    \begin{equation}
        \label{eq: residual loss}
        L_r = \frac{1}{N_c}\sum_{i=1}^{N_c}\Bigl[F\Bigl(x_i, W_N(x_i;\theta), \nabla W_N(x_i;\theta)\Bigr)\Bigr]^2.
    \end{equation}
    
    \item \textbf{Boundary Loss:}  
    At boundary points \(\{y_i\}_{i=1}^{N_b}\subset \partial D\), we enforce the boundary condition by penalizing deviations from \(\text{bc}(y_i)\):
    \begin{equation}
        \label{eq: boundary loss}
            L_b = \frac{1}{N_b}\sum_{i=1}^{N_b}\Bigl(W_N(y_i;\theta)-\text{bc}(y_i)\Bigr)^2.
    \end{equation}
    
    \item \textbf{Safe Region Loss:}  
    In a subset \(S\subset D\) where additional safe region information is available (for example, near the equilibrium for an inverted pendulum), we impose that \(W_N(x;\theta) \approx 0\). Thus,
    \begin{equation}
        \label{eq: safe loss}
            L_{\text{safe}} = \frac{1}{|S|}\sum_{x_i\in S}\Bigl(W_N(x_i;\theta)-0\Bigr)^2.
    \end{equation}
    
    \item \textbf{Unsafe Region Loss:}  
    In a subset \(U\subset D\) where points are known to be unsafe (for example, near the boundary of \(\mathcal{C}\)), we enforce that \(W_N(x;\theta) \approx 1\):
    \begin{equation}
        \label{eq: unsafe loss}
            L_{\text{unsafe}} = \frac{1}{|U|}\sum_{x_i\in U}\Bigl(W_N(x_i;\theta)-1\Bigr)^2.
    \end{equation}

\end{enumerate}

Note that the specification of $S$ and $U$ is optional, but it helps the learning process if this information is made available. The overall loss function is then given by a weighted sum:
\begin{equation}
    \label{eq: total loss}
    \mathcal{L}(\theta) = w_r\,L_r + w_b\,L_b + w_s\,L_{\text{safe}} + w_u\,L_{\text{unsafe}},
\end{equation}
where \(w_r\), \(w_b\), \(w_s\), and \(w_u\) are positive weights chosen to balance the contributions of each term.

\vspace{0.5em}
\begin{algorithm}[ht]
\caption{PINN Training for Approximating \(W_N(x;\theta)\) via the Zubov-Type PDE}
\label{algorithm: NN training}
\begin{algorithmic}[1]
\State \textbf{Input:} Interior collocation data \(S_{\text{int}}=\{x_i\}_{i=1}^{N_c} \subset D\), Boundary data \(S_{\text{b}}=\{(y_i, \text{bc}(y_i))\}_{i=1}^{N_b} \subset \partial D\), (Optional) Region-specific data for safe region \(S_{\text{safe}}\subset D\) and unsafe region \(S_{\text{unsafe}}\subset D\). Batch size \(B\), learning rate \(\eta\), maximum iterations \(M\).
\State \textbf{Output:} Trained network \(W_N(x;\theta^*)\) approximating the solution to \eqref{eq: general_pde}.
\State Initialize network parameters \(\theta\).
\State Define loss function as in \eqref{eq: total loss}.
\For{\(m=1,2,\dots, M\)}
    \State Randomly sample batches: \(S^{\text{int}}_m=\{x_i\}_{i=1}^{B}\) from \(S_{\text{int}}\), \(S^{\text{b}}_m\) from \(S_{\text{b}}\), \(S^{\text{safe}}_m\) from \(S_{\text{safe}}\), and \(S^{\text{unsafe}}_m\) from \(S_{\text{unsafe}}\).
    \State Compute \(W_N(x_i;\theta)\) and its gradient \(\nabla W_N(x_i;\theta)\) using automatic differentiation for each \(x_i\in S^{\text{int}}_m\).
    \State Evaluate the PDE residual at each \(x_i\in S^{\text{int}}_m\) using \eqref{eq: general_pde} and compute \(L_r\) as in \eqref{eq: residual loss}.
    \State Evaluate the boundary loss \(L_b\) on the batch \(S^{\text{b}}_m\) using \eqref{eq: boundary loss}.
    \State (If applicable) Evaluate the safe and unsafe region losses \(L_{\text{safe}}\) and \(L_{\text{unsafe}}\) on \(S^{\text{safe}}_m\) and \(S^{\text{unsafe}}_m\) respectively.
    \State Compute the total loss \(\mathcal{L}(\theta)\) for the current batch.
    \State Compute gradients \(\nabla_\theta \mathcal{L}(\theta)\) via backpropagation.
    \State Update parameters: \(\theta \leftarrow \theta - \eta\, \nabla_\theta \mathcal{L}(\theta)\).
    \State Compute the current loss \(\mathcal{L}_m = \mathcal{L}(\theta)\).
    \If{\(\mathcal{L}_m\) is below a specified threshold}
        \State \textbf{break}
    \EndIf
\EndFor
\State \textbf{Return:} \(\theta^*\) such that \(W_N(x;\theta^*)\) approximates the solution of \eqref{eq: general_pde}.
\end{algorithmic}
\end{algorithm}

\noindent Training is performed using standard gradient descent methods (e.g., Adam optimizer) until convergence.

\subsection{Designing a Controller from the Learned NN RBF}
\label{subsec:nn_rbf_controller}

We adopt the barrier-based control approach from Section~\ref{subsec:safe_controller}, now employing our learned neural network \(W_N(x;\theta)\) to define the barrier function. Recall from \eqref{eq:rbf_condition} that safety is enforced if the barrier condition holds. In our implementation, we first derive the barrier function \(B(x;\theta)\) from the NN output via the transformation
\begin{equation}
    \label{eq: B in terms of W}
    B(x;\theta)
=\frac{1}{2\alpha}\ln\!\Biggl(\frac{1 + W_N(x;\theta)}{1 - W_N(x;\theta)}\Biggr),
\end{equation}
as given in \eqref{eq: W in terms of B explicit} (written simply as \(B(x)\) from now on). This transformation is chosen so that the level set \(\{x\in D:W_N(x;\theta) < 1\}\) characterizes the safe set \(\mathcal{C}\) in accordance with \eqref{eq: Zubov safety condition}.

Using automatic differentiation, the gradient of \(B\) is computed as
\(
\nabla B(x)=\frac{\partial B(x)}{\partial x}.
\)
Whenever the safety condition \eqref{eq:rbf_condition} is violated (i.e., when
\[
L_fB(x)+L_gB(x)\,u_{ref}(x) - \kappa\,h(x) > 0),
\]
we apply a minimal-norm control adjustment \(\Delta u\) (as derived in \eqref{eq: delta u}) to restore safety.

The resulting control law is then given by \eqref{eq: final control} and its implementation is detailed in Algorithm~\ref{alg:barrier_controller}.

\begin{algorithm}[htbp]
\caption{Barrier-Based Controller}
\label{alg:barrier_controller}
\begin{algorithmic}[1]
\State \textbf{Input:} State \(x\), nominal control \(u_{ref}(x)\), learned NN output \(W_N(x;\theta)\), parameters \(\alpha,\kappa,\varepsilon\)
\State \textbf{Output:} Corrected control \(u(x)\)
\State Compute \(W \leftarrow W_N(x;\theta)\)
\State Compute the barrier function according to \eqref{eq: B in terms of W}.
\State Compute \(\nabla B(x)\) via automatic differentiation.
\State Evaluate Lie derivatives of \(B\) according to \eqref{eq: lie}.
\State Compute the safety measure: 
\[
s(x)= L_fB(x)+L_gB(x)u_{ref}(x)-\kappa\,h(x).
\]
\State \textbf{If} \(s(x)>0\) \textbf{then} set
\[
\Delta u=-\frac{s(x)}{\|L_gB(x)\|^2+\varepsilon}\,L_gB(x)
\]
\State \textbf{Else} set \(\Delta u=0\)
\State Output \(u(x)=u_{ref}(x)+\Delta u\)
\end{algorithmic}
\end{algorithm}

As in Section~\ref{subsec:safe_controller}, the corrected control \(u(x)\) modifies the nominal input \(u_{ref}(x)\) only when necessary to preserve safety, thus ensuring that \(\operatorname{Int}(\mathcal{C})\) remains forward invariant under the closed-loop dynamics.

\section{Experiments}
\label{section: Simulation}
In this section, we assess the efficacy of our proposed framework through three distinct case studies: an inverted pendulum system, autonomous ground navigation, and aerial navigation in obstacle-rich environments. All case studies are conducted on a computing platform equipped with an Intel  i9-11900K CPU, 32GB RAM, and NVIDIA GeForce RTX 4090 GPU.

\subsection{Inverted Pendulum}

\begin{figure}[ht]
    \centering
    \includegraphics[width=0.49\textwidth]{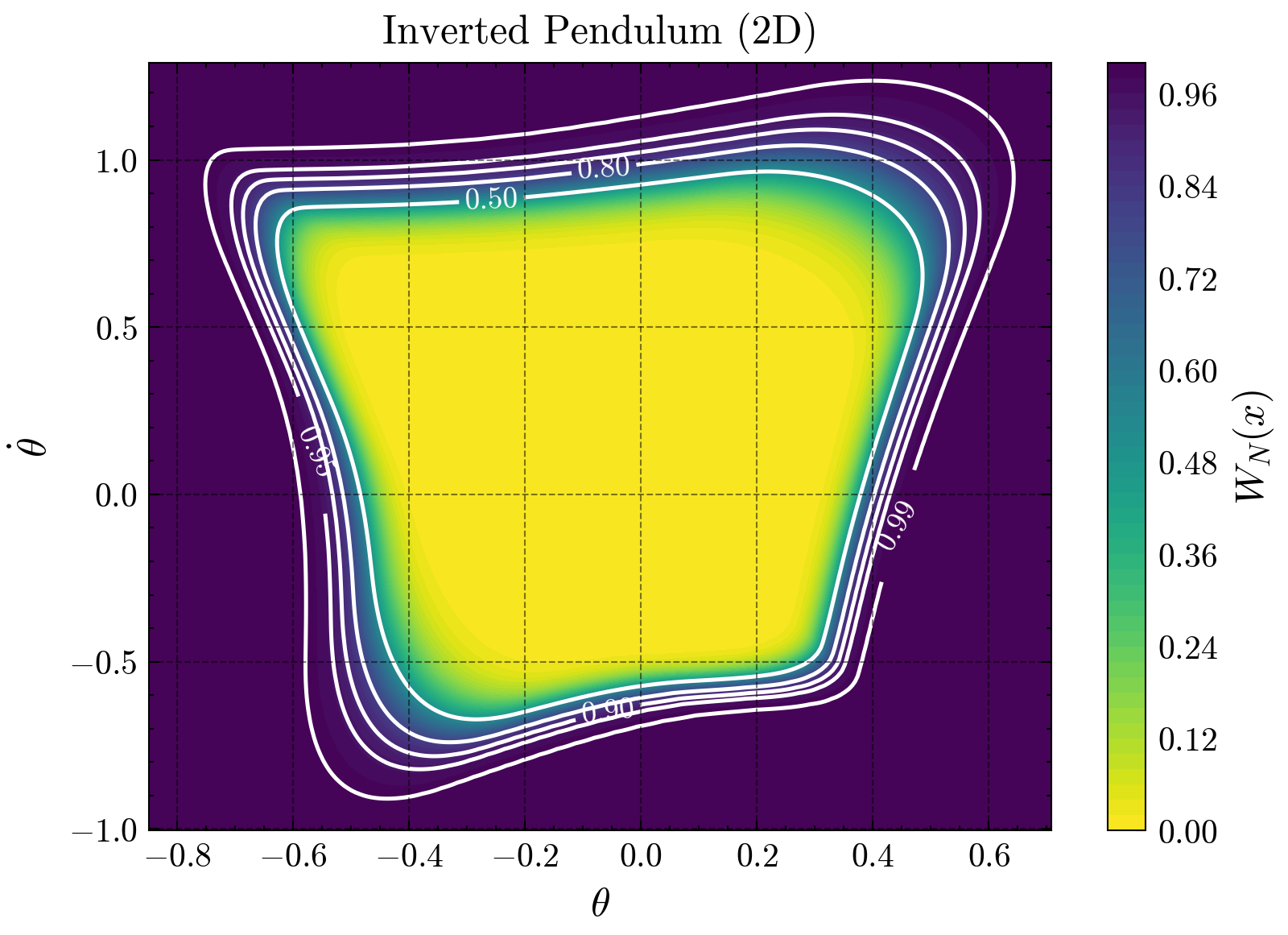}
    \vspace{-2em}
    \caption{Contour plot of the learned barrier function \(W_N(x)\) for inverted pendulum. Sub-level sets \(\{ x \in \mathbb{R}^n : W_N(x) \le \gamma \} \) for increasing \(\gamma\) illustrate how the safe region expands, as we converge to a sub-level set.}
    \vspace{-0.5em}
    \label{fig:ip_barrier}
\end{figure}

In this section, we present simulation results for an inverted pendulum example to demonstrate the proposed framework. The inverted pendulum dynamics are defined by
\begin{equation}
    \label{eq: IP dynamics}
    \dot{\theta} = \dot{\theta}, \quad \ddot{\theta} = \sin(\theta),
\end{equation}

with state \(x = [\theta,\,\dot{\theta}]^\top\). The safe region $S$ is specified as
\begin{equation}
    \label{eq: IP safe}
    |\theta| \leq \frac{\pi}{8} \quad \text{and} \quad |\dot{\theta}| \leq 1,
\end{equation}

while the unsafe region $U$ is defined by
\begin{equation}
    \label{eq: IP unsafe}
    |\theta| \geq \frac{\pi}{2} \quad \text{or} \quad |\dot{\theta}| \geq 4.
\end{equation}

A Barrier Network is trained using 10,000 random samples drawn uniformly from \(\theta \in [-\pi,\pi]\) and \(\dot{\theta} \in [-8,8]\). The training follows the Algorithm~\eqref{algorithm: NN training} for 2000 epochs. A transformed barrier function is obtained via \eqref{eq: B in terms of W} which recovers an unbounded barrier function from the bounded network output \(W_N(x)\). After training, \(W_N(x)\) is evaluated on a grid over a restricted state space, and a contour plot is generated to visualize the learned barrier function. 

As shown in Figure~\ref{fig:ip_barrier}, \(W_N(x)\) remains close to 0 in the safe region and approaches 1 near the unsafe boundaries. Sub-level sets \(\{ x \in \mathbb{R}^n : W_N(x) \le \gamma \} \) for increasing \(\gamma\) illustrate how the area enclosed by safe region also increases.

From the simulation results, we also note another potential advantage of Zubov's construction of a Barrier function, namely $W_N(x)$, that it is bounded and its value approaches one as $x$ approaches the boundary, whereas the traditional Reciprocal Barrier Functions (RBFs) approach infinity as $x$ tends to the boundary. Having a bounded function offers advantages in numerical approximation and makes it possible to extend its domain to the entire state space. By varying the threshold on $W_N (x)$, we can also adjust the effective safe region, thereby giving us flexibility in terms of choosing less (or more) restrictive safety specifications.

\subsection{Autonomous Ground Navigation}

In this section, we present simulation results for an autonomous ground navigation system using the learned neural network reciprocal barrier function (NN RBF) to enforce safety.

Our autonomous ground navigation robot is governed by the control-affine system
\begin{equation}
    \label{eq: Unicycle dynamics}
    \begin{bmatrix}
\dot{x}_1 \\
\dot{x}_2 \\
\dot{\psi}
\end{bmatrix}
=
\begin{bmatrix}
v\cos(\psi) \\
v\sin(\psi) \\
0
\end{bmatrix}
+
\begin{bmatrix}
0 \\
0 \\
1
\end{bmatrix} u,
\end{equation}
where the control influence only enters the orientation dynamics. Here, \(v>0\) is the constant forward speed, and the control input \(u\) is a scalar adjusting the robot's heading. The state \(x \in \mathbb{R}^3\), where
\(
x = [x_1 , x_2 , \psi ]^\top,
\)
is sampled uniformly from
\begin{equation}
    \label{eq: Unicycle sampling}
    x_1 \in [-2,2],\quad x_2 \in [-2,2],\quad \psi \in [-\pi,\pi].
\end{equation}

The safe region \(S\) is defined by
\begin{equation}
    \label{eq: Unicycle safe}
    |x_1| \geq 1.5 \text{ or } |x_2| \geq 1.5,
\end{equation}
while the unsafe region \(U\) is characterized as
\begin{equation}
    \label{eq: Unicycle unsafe}
    |x_1| \leq 0.2 \text{ and } |x_2| \leq 0.2.
\end{equation}

A Barrier Network is trained using 10,000 samples drawn uniformly from the above state space.

A simple reference controller is designed to steer the robot toward a desired goal (here taken as \((1,1)\)). The reference controller computes a nominal input \(u_{\text{ref}}\) based on the angular error:
\begin{equation}
    \label{eq: unicycle reference controller}
    u_{\text{ref}} = K\left(\arctan\left(\frac{1-x_2}{1-x_1}\right) - \psi\right),
\end{equation}
with gain \(K>0\).

To enforce safety, a learned NN RBF \(W_N(x;\theta)\) is obtained using the PINN training algorithm described in Algorithm~1 for 2000 epochs. The barrier-based controller (Algorithm~2) then computes a correction \(\Delta u\) by enforcing the safety condition. In this way, the barrier-based controller adjusts the nominal control to maintain safety, i.e., to ensure that the system avoids an unsafe region (here specified as a rectangular area, e.g., \([-\!0.2,0.2]\times[-0.2,0.2]\)).

\begin{figure}[ht]
    \centering
    \includegraphics[width=0.49\textwidth]{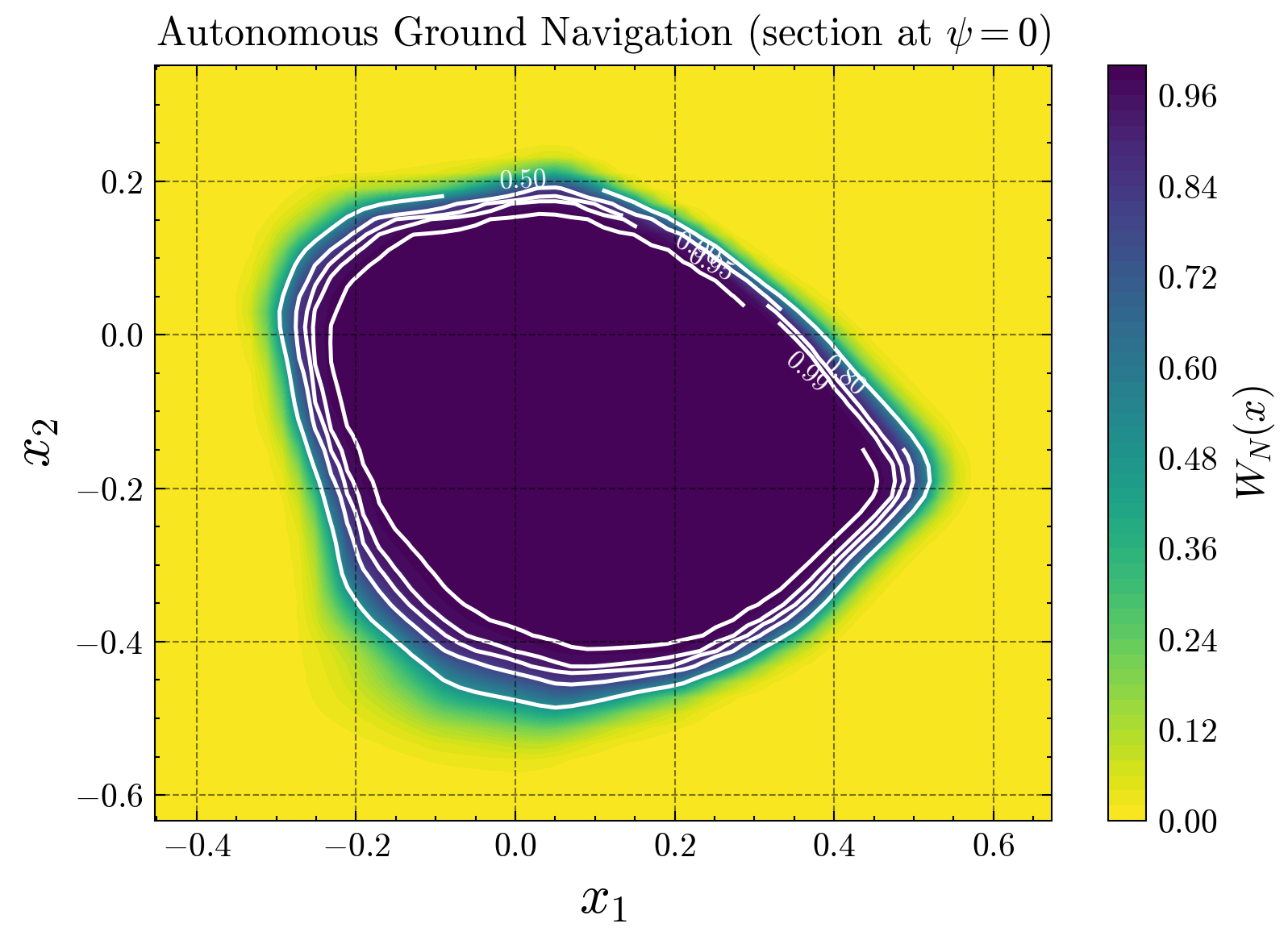}
     \vspace{-2em}
    \caption{Contour plot of \(W_N(x)\) at \(\psi = 0\) for the autonomous ground navigation system. Sub-level sets \(\{ x \in \mathbb{R}^n : W_N(x) \le \gamma \} \) for increasing \(\gamma\) illustrate how the unsafe region shrinks (leading to a larger safe area outside the contour), as we converge to a sub-level set.}
     \vspace{-0.5em}
    \label{fig:uc_barrier}
\end{figure}

Figure~\ref{fig:uc_barrier} shows that we converge to the expected CBF shape for an autonomous ground navigation robot as \(W_N(x) \rightarrow 1\). Sub-level sets \(\{ x \in \mathbb{R}^n : W_N(x) \le \gamma \} \) for increasing \(\gamma\) illustrate how the area enclosed by the contours, indicating the unsafe region in this case, decreases, showing that we are able to recover a larger safe region. This also gives us flexibility in choosing a less or more conservative barrier function by selecting the appropriate $\gamma$ value for the underlying sub-level set.

\begin{figure}[ht]
\centering
\includegraphics[width=0.49\textwidth]{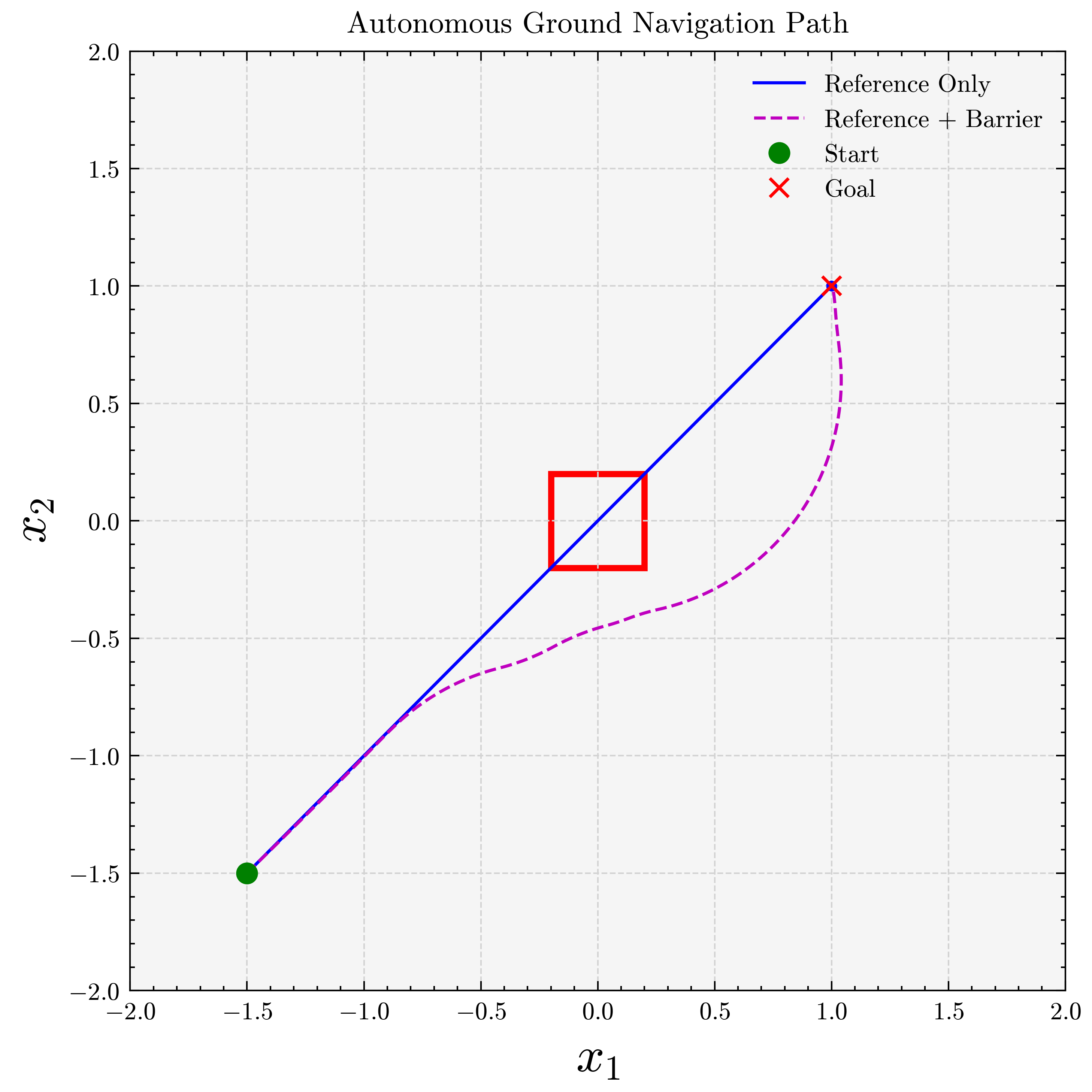}
\vspace{-2em}
\caption{Simulated autonomous ground navigation robot trajectories demonstrating obstacle avoidance when a barrier-based controller is employed.}
 \vspace{-0.5em}
\label{fig:unicycle_traj}
\end{figure}

Figure~\ref{fig:unicycle_traj} shows the simulated trajectories of the autonomous ground navigation robot. The blue trajectory corresponds to the reference controller alone, while the magenta trajectory shows the result when the barrier-based correction is applied. The unsafe region is highlighted by the red rectangle, and the start and goal positions are indicated by green and red markers, respectively. We can see that with the barrier-adjusted controller, the robot is able to avoid the obstacle/unsafe region cleanly.

\subsection{Aerial navigation in obstacle-rich environments}

In this section, we present simulation results for an aerial navigation system using a learned neural network reciprocal barrier function (NN RBF) to enhance safety. 

The aerial navigation vehicle is modeled as a six-dimensional system with state 
\begin{equation}
    \label{eq: drone state}
    x = \begin{bmatrix}
y & z & \phi & v_y & v_z & \dot{\phi}
\end{bmatrix}^\top \in \mathbb{R}^6,
\end{equation}
where \((y, z)\) denote the planar positions, \(\phi\) is the pitch angle, \((v_y, v_z)\) are the linear velocities, and \(\dot{\phi}\) is the angular velocity. The system dynamics are given by
\begin{equation}
    \label{eq: drone dynamics}
    \begin{bmatrix}
\dot{y} \\ \dot{z} \\ \dot{\phi} \\ \dot{v}_y \\ \dot{v}_z \\ \ddot{\phi}
\end{bmatrix}
=
\begin{bmatrix}
v_y \\ v_z \\ \dot{\phi} \\ 0 \\ -g \\ 0
\end{bmatrix}
+
\begin{bmatrix}
0 & 0 \\
0 & 0 \\
0 & 0 \\
-\sin\phi & -\sin\phi \\
\cos\phi & \cos\phi \\
1 & -1
\end{bmatrix} u.
\end{equation}

Here, \(g=9.81\) is the gravitational constant and the control input \(u \in \mathbb{R}^2\) corresponds to rotor speeds that are indirectly related to forces and torques acting on the vehicle.

\subsubsection{Reference Controller Design}

A PID controller is implemented as the nominal controller to steer the vehicle toward a desired target, here taken as \([1,1,0,0,0,0]^\top\). 

\subsubsection{Barrier-Based Controller Integration}

Safety is enforced by augmenting the nominal control with a barrier-based correction, as described in Algorithm~2. The learned NN RBF (obtained via Algorithm~1) is used to compute the barrier function \(B(x)\) and its gradient. 

\subsubsection{Simulation Setup and Results}

Simulations are performed using the system dynamics described above. The initial state is set to
\[
x_0 = [-1.5,\ -1.5,\ 0,\ 0,\ 0,\ 0]^\top,
\]
and the simulation is run for \(T_{\text{final}} = 10\,\text{s}\) with a time step of \(\Delta t = 0.01\,\text{s}\). Two scenarios are compared:
\begin{itemize}
    \item \textbf{Reference-Only Control:} The vehicle is controlled solely by the PID controller.
    \item \textbf{Reference + Barrier Control:} The barrier-based controller augments the nominal controller to enforce safety.
\end{itemize}

\begin{figure}[ht]
    \centering
    \includegraphics[width=0.49\textwidth]{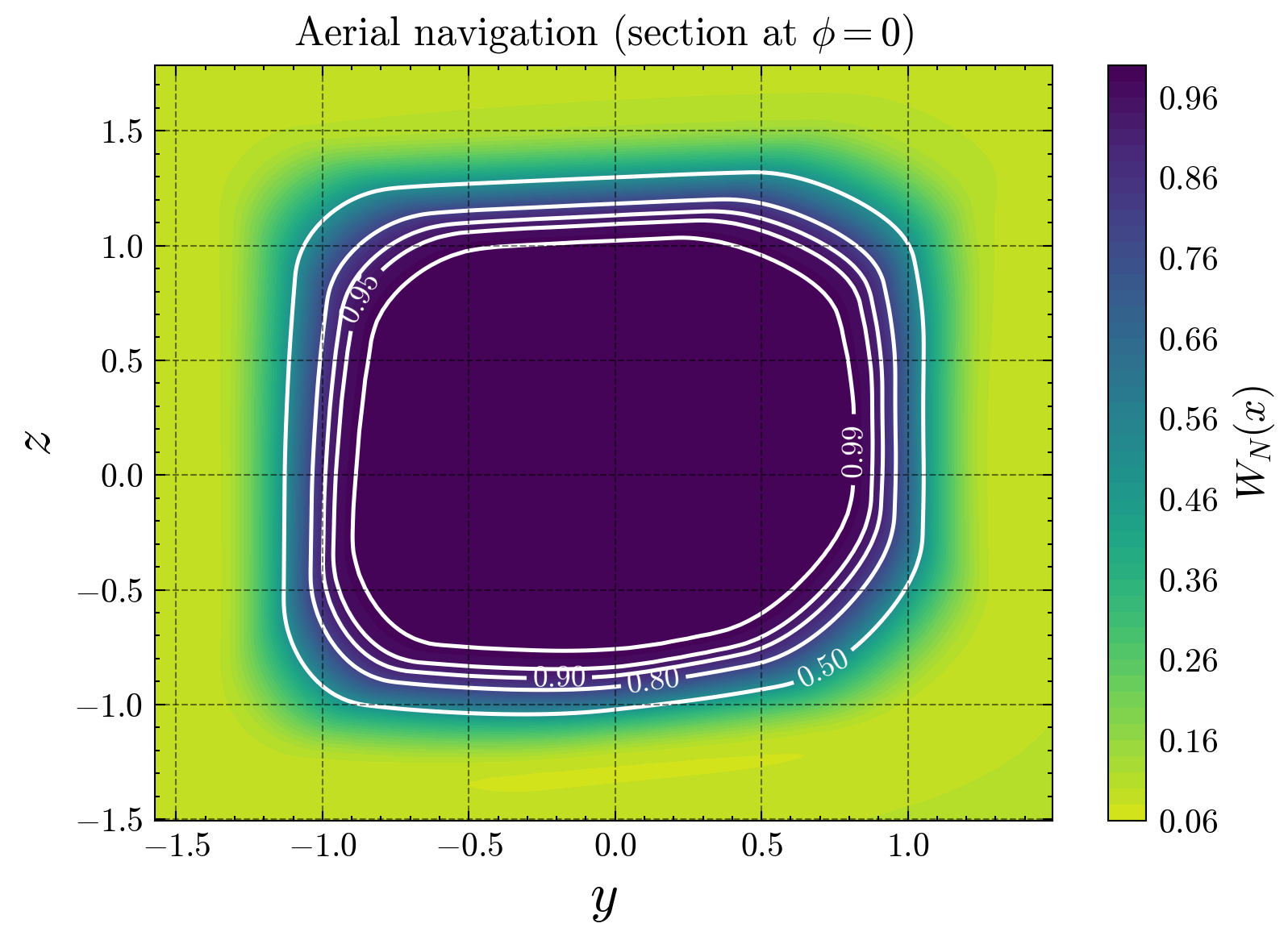}
     \vspace{-2em}
    \caption{Contour plot of \(W_N(x)\) at \(\phi = 0\) for the aerial navigation vehicle (other dimensions not shown). Sub-level sets \(\{ x \in \mathbb{R}^n : W_N(x) \le \gamma \} \) for increasing \(\gamma\) illustrate how the unsafe region shrinks (leading to a larger safe area outside the contour), as we converge to a sub-level set.}

    \label{fig:drone_barrier}
     \vspace{-0.5em}
\end{figure}

Figure~\ref{fig:drone_barrier} shows that $W_N$ converges to a well-behaved barrier shape as \(W_N(x) \rightarrow 1\). Sub-level sets \(\{ x \in \mathbb{R}^n : W_N(x) \le \gamma \} \) for increasing \(\gamma\) illustrate how the area enclosed by the contours, indicating the unsafe region, decreases, showing that we are able to recover a larger safe region. This flexibility enables us to assign an underlying $\gamma$ value of our choosing based on the safety requirements of the deployment at hand.

\begin{figure}[ht]
\centering
\includegraphics[width=0.49\textwidth]{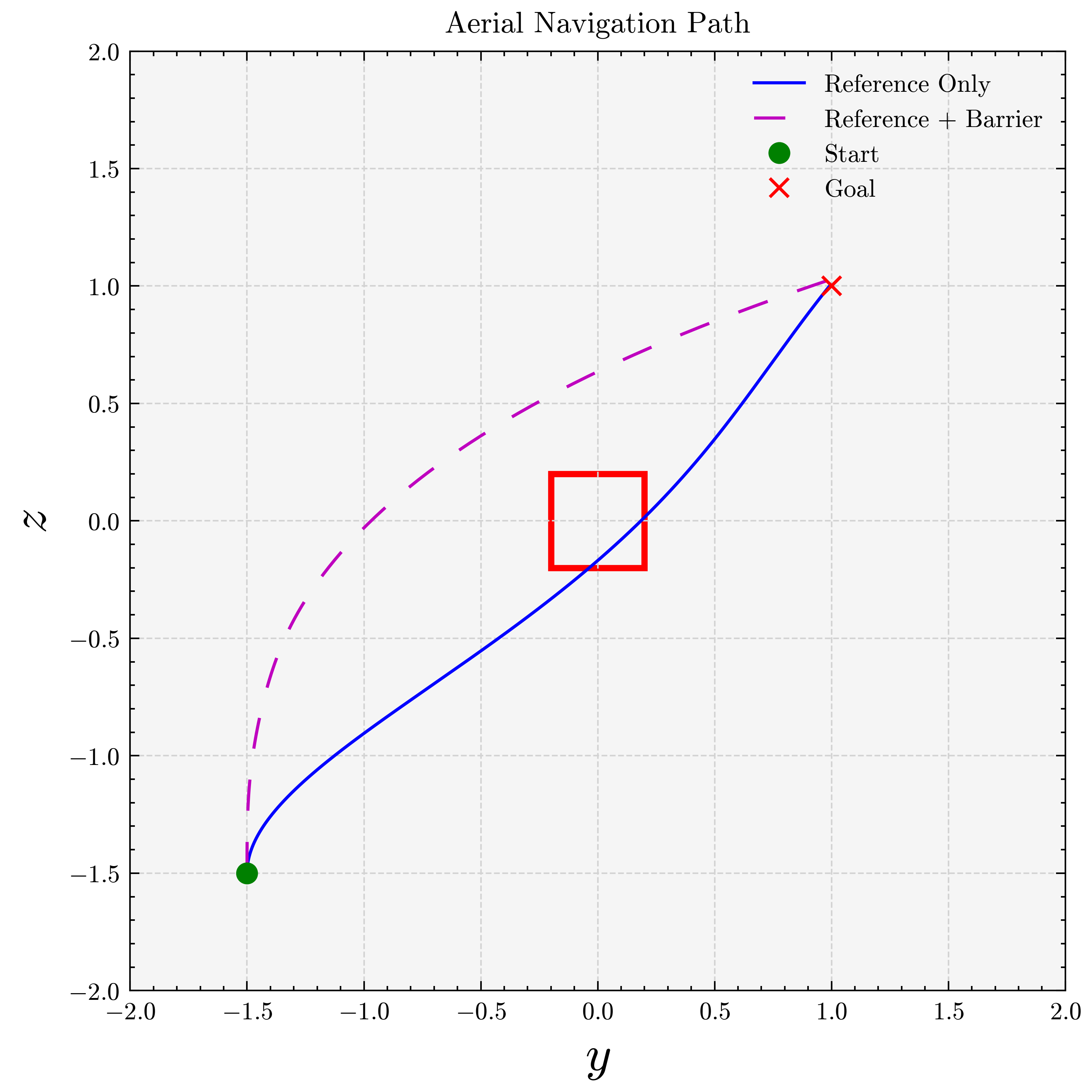}
 \vspace{-2em}
\caption{Trajectories of the aerial navigation vehicle in the \(y\)-\(z\) plane, using a trajectory following reference controller. Successful avoidance of the unsafe region is observed in the barrier-based controller due to safety adjustments to the control inputs.}
\label{fig:drone_traj}
 \vspace{-0.5em}
\end{figure}

Figure~\ref{fig:drone_traj} shows the resulting trajectories in the \(y\)-\(z\) plane. The unsafe region, defined as the rectangle, is depicted in red. The reference-only trajectory is shown in blue, while the trajectory with the barrier correction is shown in magenta. The start and target positions are marked with green and red symbols, respectively. We can clearly see that the barrier-adjusted controller effectively avoids the obstacle/unsafe region while maintaining a reasonable trajectory that reaches the goal. The result of the aerial navigation system is especially remarkable because it demonstrates efficient learning of a Control Barrier Function in a six-dimensional state space, a task many modern learning-based approaches struggle with \cite{1470374,TOPCU20082669}.

\section{Conclusion}
\label{section: Conclusions}
In this work, we presented a systematic approach to reformulate Zubov's Partial Differential Equation (PDE) within the context of safety and leveraged this formulation to synthesize neural Control Barrier Functions (CBFs) using physics-informed neural networks. Furthermore, we utilized the learned neural CBFs to design a safe controller through a Quadratic Program (QP). A key advantage of our framework lies in its flexibility, allowing the definition of custom level sets that enable users to adapt the safe region according to specific safety requirements. To evaluate the efficacy of our approach, we conducted experiments on various control tasks, including an inverted pendulum, autonomous ground navigation, and aerial navigation in obstacle-laden environments. The results demonstrate that our method effectively scales to high-dimensional systems while ensuring safety.

Future work may explore:
\begin{itemize}
    \item Modifying the loss function under the proposed training scheme to account for multiple constraints by learning a single \(W_N\) function instead of composing multiple barrier functions \cite{aali2022multiplecontrolbarrierfunctions,breeden2023compositionsmultiplecontrolbarrier}.
    \item Extending our approach from Reciprocal Barrier Functions (RBFs) to Zeroing Barrier Functions (ZBFs) \cite{Ames_2017, ames2019control}.
    \item Extensions to more complex systems and incorporating robustness against model uncertainties and verification schemes to provide formal guarantees on the learned Neural CBF.
\end{itemize}

\label{section: References}
\bibliographystyle{IEEEtran}
\bibliography{references.bib}

\begin{thebibliography}{10}
\providecommand{\url}[1]{#1}
\csname url@samestyle\endcsname
\providecommand{\newblock}{\relax}
\providecommand{\bibinfo}[2]{#2}
\providecommand{\BIBentrySTDinterwordspacing}{\spaceskip=0pt\relax}
\providecommand{\BIBentryALTinterwordstretchfactor}{4}
\providecommand{\BIBentryALTinterwordspacing}{\spaceskip=\fontdimen2\font plus
\BIBentryALTinterwordstretchfactor\fontdimen3\font minus \fontdimen4\font\relax}
\providecommand{\BIBforeignlanguage}[2]{{%
\expandafter\ifx\csname l@#1\endcsname\relax
\typeout{** WARNING: IEEEtran.bst: No hyphenation pattern has been}%
\typeout{** loaded for the language `#1'. Using the pattern for}%
\typeout{** the default language instead.}%
\else
\language=\csname l@#1\endcsname
\fi
#2}}
\providecommand{\BIBdecl}{\relax}
\BIBdecl

\bibitem{altman2021constrained}
E.~Altman, \emph{Constrained Markov decision processes}.\hskip 1em plus 0.5em minus 0.4em\relax Routledge, 2021.

\bibitem{achiam2017constrained}
J.~Achiam, D.~Held, A.~Tamar, and P.~Abbeel, ``Constrained policy optimization,'' in \emph{International conference on machine learning}.\hskip 1em plus 0.5em minus 0.4em\relax PMLR, 2017, pp. 22--31.

\bibitem{8263977}
S.~Bansal, M.~Chen, S.~Herbert, and C.~J. Tomlin, ``Hamilton-{J}acobi reachability: {A} brief overview and recent advances,'' in \emph{2017 IEEE 56th Annual Conference on Decision and Control (CDC)}, 2017.

\bibitem{tayal2025physics}
M.~Tayal, A.~Singh, S.~Kolathaya, and S.~Bansal, ``A physics-informed machine learning framework for safe and optimal control of autonomous systems,'' \emph{arXiv preprint arXiv:2502.11057}, 2025.

\bibitem{singh2024exactimpositionsafetyboundary}
\BIBentryALTinterwordspacing
A.~Singh, Z.~Feng, and S.~Bansal, ``Exact imposition of safety boundary conditions in neural reachable tubes,'' 2024. [Online]. Available: \url{https://arxiv.org/abs/2404.00814}
\BIBentrySTDinterwordspacing

\bibitem{ames2014control}
A.~D. Ames, J.~W. Grizzle, and P.~Tabuada, ``Control barrier function based quadratic programs with application to adaptive cruise control,'' in \emph{53rd IEEE Conference on Decision and Control}.\hskip 1em plus 0.5em minus 0.4em\relax IEEE, 2014, pp. 6271--6278.

\bibitem{Ames_2017}
A.~D. Ames, X.~Xu, J.~W. Grizzle, and P.~Tabuada, ``Control barrier function based quadratic programs for safety critical systems,'' \emph{{IEEE} Transactions on Automatic Control}, vol.~62, no.~8, pp. 3861--3876, 2017.

\bibitem{7525253}
G.~Wu and K.~Sreenath, ``Safety-critical control of a planar quadrotor,'' in \emph{2016 American Control Conference (ACC)}, 2016, pp. 2252--2258.

\bibitem{tayal2024control}
M.~Tayal, R.~Singh, J.~Keshavan, and S.~Kolathaya, ``Control barrier functions in dynamic uavs for kinematic obstacle avoidance: A collision cone approach,'' in \emph{2024 American Control Conference (ACC)}.\hskip 1em plus 0.5em minus 0.4em\relax IEEE, 2024, pp. 3722--3727.

\bibitem{ames2019control}
A.~D. Ames, S.~Coogan, M.~Egerstedt, G.~Notomista, K.~Sreenath, and P.~Tabuada, ``Control barrier functions: Theory and applications,'' in \emph{18th European control conference (ECC)}.\hskip 1em plus 0.5em minus 0.4em\relax IEEE, 2019, pp. 3420--3431.

\bibitem{C3BF-Legged}
M.~Tayal and S.~Kolathaya, ``Safe legged locomotion using collision cone control barrier functions (c3bfs),'' \emph{arXiv preprint arXiv:2309.01898}, 2023.

\bibitem{1470374}
A.~Papachristodoulou and S.~Prajna, ``A tutorial on sum of squares techniques for systems analysis,'' in \emph{Proceedings of the 2005, American Control Conference, 2005.}, 2005, pp. 2686--2700 vol. 4.

\bibitem{TOPCU20082669}
U.~Topcu, A.~Packard, and P.~Seiler, ``Local stability analysis using simulations and sum-of-squares programming,'' \emph{Automatica}, vol.~44, no.~10, pp. 2669--2675, 2008.

\bibitem{9303785}
A.~Robey, H.~Hu, L.~Lindemann, H.~Zhang, D.~V. Dimarogonas, S.~Tu, and N.~Matni, ``Learning control barrier functions from expert demonstrations,'' in \emph{2020 59th IEEE Conference on Decision and Control (CDC)}, 2020, pp. 3717--3724.

\bibitem{zhao2020synthesizing}
H.~Zhao, X.~Zeng, T.~Chen, and Z.~Liu, ``Synthesizing barrier certificates using neural networks,'' in \emph{Proceedings of the 23rd international conference on hybrid systems: Computation and control}, 2020, pp. 1--11.

\bibitem{abate2021fossil}
A.~Abate, D.~Ahmed, A.~Edwards, M.~Giacobbe, and A.~Peruffo, ``Fossil: A software tool for the formal synthesis of {L}yapunov functions and barrier certificates using neural networks,'' in \emph{Proceedings of the 24th International Conference on Hybrid Systems: Computation and Control}, 2021, pp. 1--11.

\bibitem{zhao2022verifying}
Q.~Zhao, X.~Chen, Z.~Zhao, Y.~Zhang, E.~Tang, and X.~Li, ``Verifying neural network controlled systems using neural networks,'' in \emph{25th ACM International Conference on Hybrid Systems: Computation and Control}, 2022, pp. 1--11.

\bibitem{NEURIPS2023_120ed726}
H.~Zhang, J.~Wu, Y.~Vorobeychik, and A.~Clark, ``Exact verification of relu neural control barrier functions,'' in \emph{Advances in Neural Information Processing Systems}, A.~Oh, T.~Neumann, A.~Globerson, K.~Saenko, M.~Hardt, and S.~Levine, Eds., vol.~36.\hskip 1em plus 0.5em minus 0.4em\relax Curran Associates, Inc., 2023, pp. 5685--5705.

\bibitem{dawson2023safe}
C.~Dawson, S.~Gao, and C.~Fan, ``Safe control with learned certificates: A survey of neural {L}yapunov, barrier, and contraction methods for robotics and control,'' \emph{IEEE Transactions on Robotics}, 2023.

\bibitem{liu2023safe}
S.~Liu, C.~Liu, and J.~Dolan, ``Safe control under input limits with neural control barrier functions,'' in \emph{Conference on Robot Learning}.\hskip 1em plus 0.5em minus 0.4em\relax PMLR, 2023, pp. 1970--1980.

\bibitem{tayal2025cp}
M.~Tayal, A.~Singh, P.~Jagtap, and S.~Kolathaya, ``Cp-ncbf: A conformal prediction-based approach to synthesize verified neural control barrier functions,'' \emph{arXiv preprint arXiv:2503.17395}, 2025.

\bibitem{9993334}
B.~Dai, P.~Krishnamurthy, and F.~Khorrami, ``Learning a better control barrier function,'' in \emph{2022 IEEE 61st Conference on Decision and Control (CDC)}, 2022, pp. 945--950.

\bibitem{tayal2024learning}
M.~Tayal, H.~Zhang, P.~Jagtap, A.~Clark, and S.~Kolathaya, ``Learning a formally verified control barrier function in stochastic environment,'' in \emph{2024 IEEE 63rd Conference on Decision and Control (CDC)}, 2024, pp. 4098--4104.

\bibitem{tayal2024semi}
M.~Tayal, A.~Singh, P.~Jagtap, and S.~Kolathaya, ``Semi-supervised safe visuomotor policy synthesis using barrier certificates,'' \emph{arXiv preprint arXiv:2409.12616}, 2024.

\bibitem{clf-zubov}
F.~Camilli, L.~Grüne, and F.~Wirth, ``Control lyapunov functions and zubov's method,'' \emph{SIAM J. Control and Optimization}, vol.~47, pp. 301--326, 01 2008.

\bibitem{liu2023learningverifyingmaximalneural}
\BIBentryALTinterwordspacing
J.~Liu, Y.~Meng, M.~Fitzsimmons, and R.~Zhou, ``Towards learning and verifying maximal neural lyapunov functions,'' 2023. [Online]. Available: \url{https://arxiv.org/abs/2304.07215}
\BIBentrySTDinterwordspacing

\bibitem{liu2024formallyverifiedphysicsinformedneural}
\BIBentryALTinterwordspacing
J.~Liu, M.~Fitzsimmons, R.~Zhou, and Y.~Meng, ``Formally verified physics-informed neural control lyapunov functions,'' 2024. [Online]. Available: \url{https://arxiv.org/abs/2409.20528}
\BIBentrySTDinterwordspacing

\bibitem{liu2025physicsinformedneuralnetworklyapunov}
\BIBentryALTinterwordspacing
J.~Liu, Y.~Meng, M.~Fitzsimmons, and R.~Zhou, ``Physics-informed neural network lyapunov functions: Pde characterization, learning, and verification,'' 2025. [Online]. Available: \url{https://arxiv.org/abs/2312.09131}
\BIBentrySTDinterwordspacing

\bibitem{zubov1964methods}
\BIBentryALTinterwordspacing
V.~I. Zubov, ``\BIBforeignlanguage{English}{Methods of a. m. lyapunov and their applications},'' 1964, retrieved from Archive.org on 2025-04-01. [Online]. Available: \url{https://archive.org/details/v.-i.-zubov-methods-of-a.-m.-lyapunov-and-their-applications-1964}
\BIBentrySTDinterwordspacing

\bibitem{converse}
J.~Liu, ``Converse barrier functions via lyapunov functions,'' \emph{IEEE Transactions on Automatic Control}, vol.~67, no.~1, pp. 497--503, 2022.

\bibitem{aali2022multiplecontrolbarrierfunctions}
\BIBentryALTinterwordspacing
M.~Aali and J.~Liu, ``Multiple control barrier functions: An application to reactive obstacle avoidance for a multi-steering tractor-trailer system,'' 2022. [Online]. Available: \url{https://arxiv.org/abs/2209.05156}
\BIBentrySTDinterwordspacing

\bibitem{breeden2023compositionsmultiplecontrolbarrier}
\BIBentryALTinterwordspacing
J.~Breeden and D.~Panagou, ``Compositions of multiple control barrier functions under input constraints,'' 2023. [Online]. Available: \url{https://arxiv.org/abs/2210.01354}
\BIBentrySTDinterwordspacing

\end{thebibliography}

\end{document}